# Layout-Independent License Plate Recognition via Integrated Vision and Language Models

Elham, Shabaninia*

Department of Applied Mathematics, Graduate University of Advanced Technology, Kerman, Iran,

e.shabaninia@kgut.ac.ir

Fatemeh Asadi-zeydabadi

Intelligent Data Processing Laboratory, Department of Electrical Engineering, Shahid Bahonar University of Kerman, Kerman, Iran,

fatemeh.asadi@eng.uk.ac.ir

Hossein, Nezamabadi-pour

Intelligent Data Processing Laboratory, Department of Electrical Engineering, Shahid Bahonar University of Kerman, Kerman, Iran,

nezam@uk.ac.ir

## Abstract

This work presents a pattern-aware framework for automatic license plate recognition (ALPR), designed to operate reliably across diverse plate layouts and challenging real-world conditions. The proposed system consists of a modern, high-precision detection network followed by a recognition stage that integrates a transformer-based vision model with an iterative language modelling mechanism. This unified recognition stage performs character identification and post-OCR refinement in a seamless process, learning the structural patterns and formatting rules specific to license plates without relying on explicit heuristic corrections or manual layout classification. Through this design, the system jointly optimizes visual and linguistic cues, enables iterative refinement to improve OCR accuracy under noise, distortion, and unconventional fonts, and achieves layout-independent recognition across multiple international datasets (IR-LPR, UFPR-ALPR, AOLP). Experimental results demonstrate superior accuracy and robustness compared to recent segmentation-free approaches, highlighting how embedding pattern analysis within the recognition stage bridges computer vision and language modelling for enhanced adaptability in intelligent transportation and surveillance applications.

---

* Corresponding author.

# 1. Introduction

Intelligent Transportation Systems (ITS) have become indispensable tools for municipalities and governments in implementing modern traffic management strategies. With the increasing number of vehicles on the road, managing traffic efficiently is challenging, even in smaller areas like public parking lots, if solely relying on human personnel. To address these challenges, ITS are typically integrated with video surveillance and enforcement systems, using cameras strategically placed along roadways and in urban environments [1].

A crucial component of ITS is Automatic License Plate Recognition (ALPR), which leverages machine vision and artificial intelligence to recognize and process vehicle license plates. As a unique vehicle identifier, the license plate enables a wide range of ALPR applications, including automatic toll collection on highways [2, 3], parking management, border control, traffic monitoring [4, 5], tracking stolen vehicles, identifying traffic violations [6, 7], and regulating access to secure areas. The accurate extraction of characters from license plates is vital for subsequent administrative actions and legal enforcement.

A typical ALPR system comprises three core modules: License Plate Detection (LPD), Character Segmentation (CS), and Optical Character Recognition (OCR). The LPD module first identifies and isolates the region containing the license plate for further processing. The CS module then segments the characters within the plate area, preparing them for the final phase, OCR, where each character is classified and recognized. Although this multi-step process is well-established in the literature [6-10], it faces significant challenges, particularly in the segmentation and annotation phases, due to the character-level nature of the recognition task. These challenges are exacerbated in uncontrolled environments, where factors such as occlusion, uneven lighting, rotation, and blurriness can severely impact accuracy. Recent advancements in segmentation-free approaches for license plate recognition have shown promising results [11-14]. However, despite improvements in accuracy and efficiency, these solutions still struggle with robustness in real-world applications.

Moreover, ALPR systems often require post-OCR correction to verify specific license plate patterns based on layout. The diversity of license plate formats and languages across different countries further complicates the development of adaptable ALPR systems. For example, in the United States, each state has its own design, character set, and numbering system, such as California's "1ABC234" and Texas's "ABC-1234" . The United Kingdom uses a format that indicates the vehicle's age and location, with distinct front (white) and rear (yellow) backgrounds. These variations demand ALPR systems capable of recognizing a wide array of patterns and characters, as illustrated in Figure 1.

In contrast, some contemporary optical character recognition (OCR) approaches utilize end-to-end networks that seamlessly combine text recognition and post-OCR correction into a unified process [15]. By incorporating language modeling, contextual understanding of the text is ensured, which not only enhances the accuracy of the recognized characters but also improves the efficiency of the overall process by reducing the need for separate correction stages. This unified approach exemplifies the shift towards more sophisticated and streamlined OCR systems in modern research.

Building on this foundation, we propose a pattern‑aware vision–language recognition framework that directly addresses these limitations. A modern, high‑precision object detector first localises the license plate region, which is then processed by an integrated recognition stage combining a convolutional transformer‑based vision model with an iterative language modelling mechanism. This unified stage performs character recognition and output refinement in a single pipeline, enabling the system to automatically learn the structural patterns and formatting constraints of license plates from data. By avoiding reliance on rule‑based post‑processing, the framework can be adapted to new plate layouts simply through retraining with relevant images, without the need for manual format specification. This work extends our earlier study focused solely on the recognition stage of license plate reading [16], by generalising the approach to multiple international datasets and enhancing robustness under diverse real-world conditions. The key contributions of this study are:

1. **Layout-independent recognition architecture** that embeds structural pattern analysis within the recognition process.

2. **Iterative refinement mechanism** leveraging visual–linguistic cues to enhance OCR results under challenging conditions.

3. **Cross-dataset validation** on IR-LPR, UFPR-ALPR, and AOLP, proving scalability across different languages, fonts, and designs.

4. **Segmentation-free operation** that removes a traditional ALPR bottleneck while improving precision and robustness.

The remainder of this paper is organized as follows: Section 2 reviews related approaches to automatic license plate recognition. Section 3 describes the proposed method, including the license plate detection and recognition network. Section 4 presents the evaluation results, and Section 5 concludes the paper with insights and future research directions.

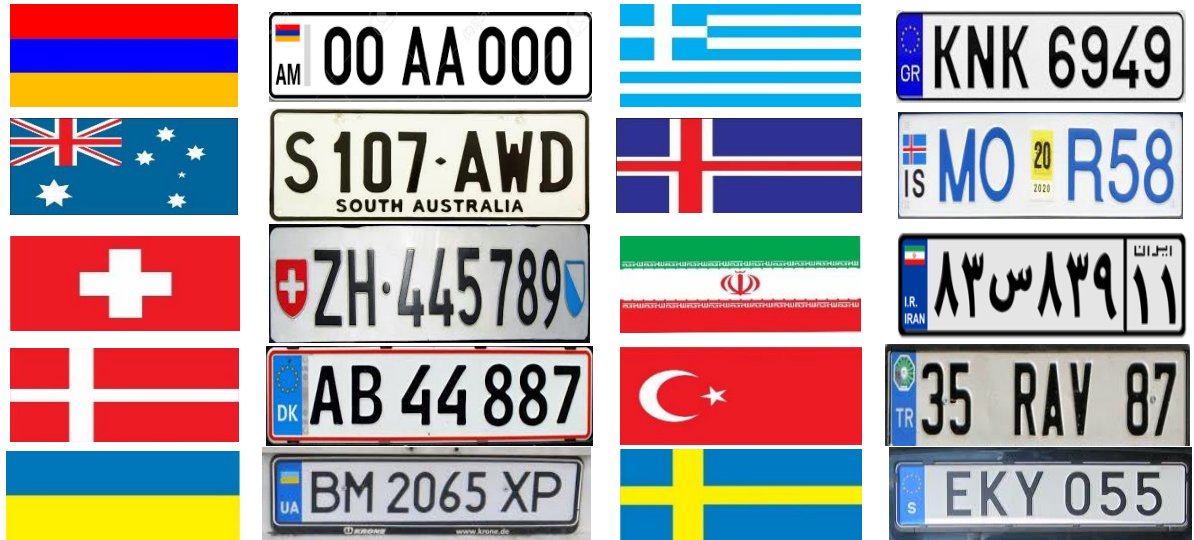

**Fig. 1.** Some examples for visual representation of global diversity in license plate formats and languages.

## 2. Background and Related Work

With the rapid advancement of deep learning technology, the field of ALPR has experienced significant development opportunities due to the robust feature learning and generalization capabilities of deep learning models. This has led to increased efforts to address unconstrained scenarios, thereby expanding the applicability of ALPR systems. This section reviews and summarizes relevant research on deep learning-based approaches for the two essential modules of ALPR systems: detection and recognition.

### 2.1. License Plate Detection

License plates can be identified using a two-step process: first, detecting the vehicles, and then detecting the license plates [17]. Although this method can improve accuracy, it typically introduces additional time overhead and is not commonly used. In recent years, efficient algorithms have emerged in the field of computer vision for recognizing and identifying license plates. Some of these approaches are free from traditional object detectors, focusing instead on specialized techniques for challenging scenarios. In [18], a hybrid cascade structure is specifically designed to detect small and blurred license plates in complex scenes, bypassing the need for conventional object detection frameworks. Similarly, in [19], a Recurrent Neural Network (RNN) is proposed to enhance the localization accuracy of license plates in difficult environments. Ref. [20] improves detection by predicting the four corners of a license plate, helping to better isolate it in complex images.

In contrast, many effective methods rely on deep learning-based object detectors, which have proven highly accurate in near real-time applications. For example, Faster R-CNN has been widely used for vehicle and license plate identification across diverse environments [21]. Such methods are generally robust, benefiting from extensive training datasets. For directly locating license plates, well-known object detection models like YOLO [22], SSD [23], and Faster R-CNN [24] are also commonly applied. YOLO (You Only Look Once), introduced by Redmon et al. in 2016, directly predicts bounding boxes and class probabilities with a single neural network. YOLO's simplicity allows for rapid predictions [25], where an image is divided into an S×S grid, and each cell predicts B bounding boxes along with confidence scores. The confidence score reflects the likelihood of an object being present, based on the Intersection over Union (IoU) between predicted boxes and ground truth. Researchers have widely adopted and modified YOLO to improve its performance in specific applications [26, 27]. For instance, Reference [28] introduces a license plate recognition system utilizing the YOLO detector, while Reference [29] presents a multi-directional detector based on a modified YOLO framework specifically designed for license plate images.

In conclusion, YOLO is a highly efficient and accurate method for license plate detection, striking an excellent balance between speed and precision. Its single-step prediction and rapid processing capabilities make it particularly effective for real-time applications, while its flexibility and adaptability to specific scenarios further solidify its suitability for diverse license plate detection tasks.

### 2.2. License Plate Recognition (LP Recognition)

The last step is to recognize the license plate characters. Traditional character recognition methods are typically based on segmentation. Thanks to the development of deep learning technology and hardware computing power in recent years, segmentation-free methods have gradually become mainstream. Before character segmentation or recognition, it is usually necessary to carry out preprocessing operations, such as affine and perspective transformation, to deal with license plate rotation, tilt, and deformation [30].

### 2.2.1. Segmentation-Based Methods

Character segmentation is based on the truth that characters and background have obviously different colors in a license plate. Usually, the binary image of a license plate is needed to obtain the boundary of characters by horizontal pixel projection [31-34]. Then, the problem is transformed into character recognition, which is a kind of image classification task. It should be noted that the accuracy of license plate recognition depends heavily on the quality of character segmentation. The feature extractor [35] extracts the salient features of characters from the image to be recognized without all pixels of a specific character, which reduces the calculation cost and is robust to character rotation and noise. The deep learning-based methods [34] can directly obtain the character features from the original pixel data, equivalent to acting as a feature extractor and classifier.

### 2.2.2. Segmentation-Free Methods

In order to avoid the uncertainty of character segmentation, segmentation-free methods have become a hot research topic in recent years. Segmentation-free methods usually take the license plate characters as a sequence directly. The recognition task can be transformed into sequence labeling problem. To solve this type of problem, both CNN-based and RNN-based methods have been adopted in different situations. Wang et al. [36] used a bi-directional recurrent neural network (BRNN) and Connectionist Temporal Classification (CTC) structure to identify the license plate characters from end to end. Reference [37] divided the character recognition model into two parts. The first part uses CNN to extract image features, and the second part uses LSTM and CTC to obtain context information. Some studies [38-41] also use object detection methods to transform the character classification problem into an object recognition problem, using a lightweight and fast object detection method to detect characters on the detected license plate area.

license plate recognition is typically followed by the application of heuristic rules to adjust results based on the expected layout format specific to a country or state. This post-processing technique helps ensure the correct number of letters and digits and reduces errors in frequently misclassified characters, such as 'G' and '6', and 'I' and '1', among others. In this context, while a tailored ALPR system can handle specific license plate layouts within a region, it requires verification or modification when new LP layouts are introduced or when recognizing plates from neighboring regions.

In contrast, this paper proposes a segmentation-free approach that also learns the language of license plates during recognition and simplifies adaptation by only requiring retraining of the network with images of the new layout. This method does not require any layout classification or heuristic rules, and allows for easy adjustments to accommodate different or additional LP layouts.

## 3. Proposed ALPR System

This section outlines our proposed method for automatic license plate recognition, which is divided into two stages. The first stage focuses on detecting the license plates, while the second stage handles character recognition on the detected plates. A schematic overview of our proposed framework is presented in Figure 2.

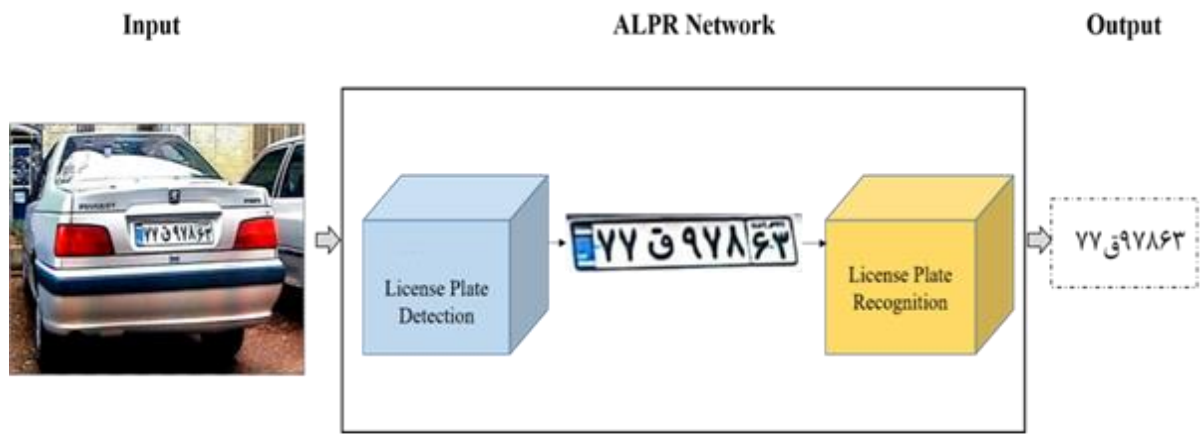

**Fig. 2.** A schematic representation of our proposed framework.

### 3.1. License Plate Detection Network

The advancements in object detection achieved by YOLO have had a significant impact on ALPR systems. The YOLOv9 architecture [42] builds on its predecessors with several key enhancements aim at improving both accuracy and efficiency. Here's an overview of its main components:

**Enhanced Backbone Network:** Employs an advanced convolutional neural network architecture, potentially a variant of CSPDarknet [43] or another optimized backbone, for superior feature extraction.

**Improved Detection Head:** Refined to enhance the precision and recall of bounding boxes, featuring better anchor box handling and more accurate class predictions.

**Path Aggregation Network (PANet):** Advances the feature pyramid network (FPN) [44] by facilitating improved information flow between layers, thereby enhancing object detection across various scales.

**Advanced Post-Processing:** Utilizes sophisticated techniques like Non-Maximum Suppression (NMS) [45] and optimized Intersection over Union (IoU) thresholds to minimize false positives and improve detection quality.

**Enhanced Data Augmentation and Training Techniques:** Incorporates advanced strategies such as mosaic and mixup [46] for data augmentation, alongside robust training techniques like label smoothing and intelligent loss functions, to further boost performance.

YOLOv9 is designed for scalability, making it adaptable to different hardware capabilities and performance requirements, from high-end GPUs to edge devices. It maintains the real-time detection capabilities of YOLO models while offering notable improvements in accuracy and efficiency, making it suitable for a broad range of applications.

### 3.2. License Plate Recognition Network

As mentioned before, the method used for recognition is taken from [15], which includes two units of a convolutional Transformer-based vision mode (VM) and a language model (LM). This architecture is illustrated in Figure 3.

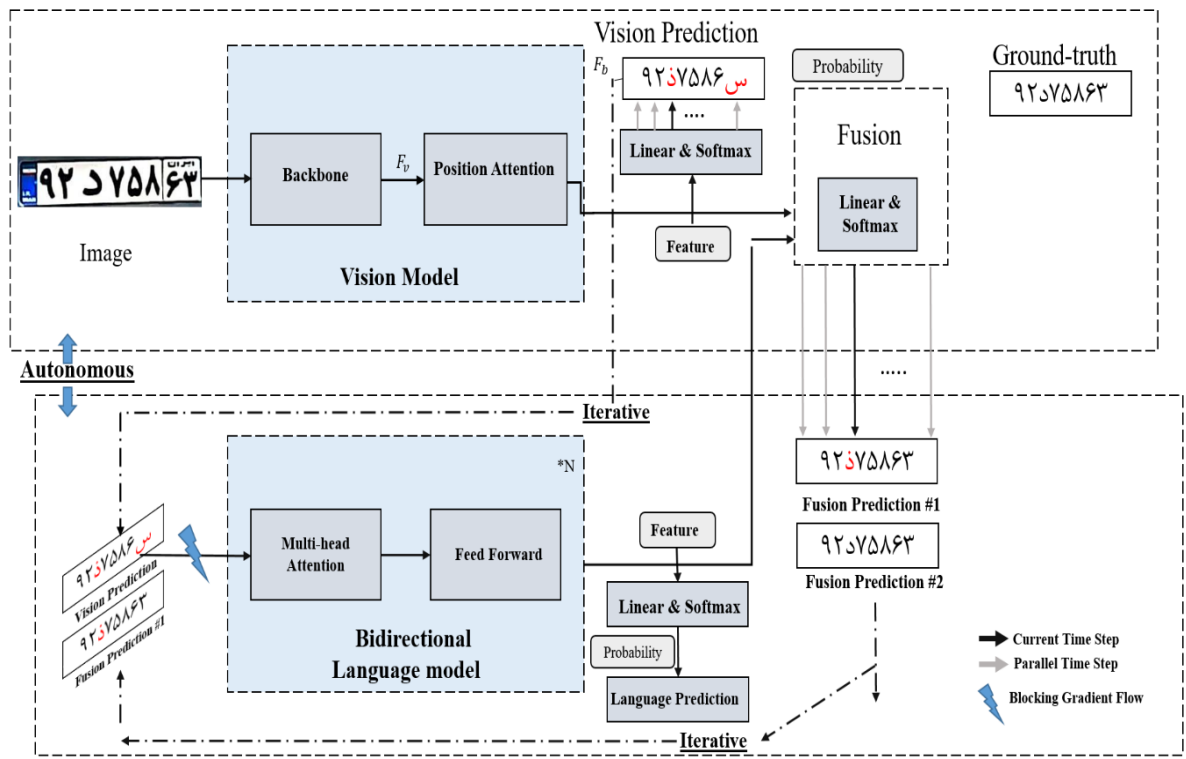

**Fig. 3.** An illustration of the LP recognition method. Image courtesy of [15].

### 3.2.1. Vision Model (VM)

The VM utilizes a Convolutional Transformer-based (CvT) architecture with a Transformer-based position attention block. It incorporates a ResNet45 [47] convolutional backbone (B) for initial feature extraction ($F_b$), followed by Transformer for sequence modeling (M) [48, 49]. This approach integrates global context and dynamic attention mechanisms to enhance performance while optimizing computational resources. It works by taking as input the x image with the width w and height h.

$$F_b = \mathcal{B}(x) \in R^{h \times w \times d} \quad (1)$$

$$F_m = M(F_b) \in R^{h \times w \times d} \quad (2)$$

In the query paradigm, the position attention unit converts visual features into character probabilities [48] in parallel:

$$Q = PE(t) \in R^{h \times w \times d} \quad (3)$$

$$K = g(F_m) \, R^{h \times w \times d} \quad (4)$$

$$V = H(F_m) \, R^{h \times w \times d} \quad (5)$$

$$F_v = Softmax\left(\frac{QK^T}{\sqrt{D}}\right)V \quad (6)$$

where $Q$ represents positional encodings (PE) for character order, t is the character sequence length, $K$ is derived from $g(F_b)$ similar to [50] that is implemented by a U-net like network, and $V$ comes from $H(F_b)$, with $H(.)$ being an identity mapping.

### 3.2.2. Language Model (LM)

In [15], a bidirectional cloze network (BCN) is employed as the language model, which is essentially a modified version of an L-layer Transformer decoder. Each layer in BCN comprises cross-attention multi-head attention mechanisms and feed-forward networks, followed by residual connections and layer normalization. Unlike the standard Transformer, BCN directly feeds character vectors into the multi-head attention blocks, bypassing the first layer of the network. Moreover, the attention mask within the multi-head attention is designed to prevent the model from accessing the same timestep, thereby avoiding self-reference. BCN also omits self-attention to prevent information leakage across different timesteps.

This strategy effectively treats the LM as an independent spelling correction system, where it processes character probability vectors to produce character probability distributions. BCN improves efficiency by utilizing attention masks to prevent self-attention and information leakage. It integrates Transformer decoder layers, with character vectors being directly input into the multi-head attention blocks. The attention mechanism in BCN is defined as:

$$M_{ij} = \begin{cases} 0, & i \neq j \\ -\infty, & i = j \end{cases}$$
$$K_i = V_i = P(y_i)W\iota$$
$$F_{mha} = Softmax\left(\frac{QK^T}{\sqrt{C}} + M\right)V \quad (7)$$

where $Q$ represents positional encodings or outputs from the last layer, $K$ and $V$ are derived from the character probability $P(y_i)$ For a text string $y = (y_1, ..., y_n)$, and $M$ is the attention mask matrix. BCN's Transformer-like architecture allows for independent and parallel computations, requiring half the parameters and computations compared to aggregate models. To handle noisy inputs, the language model is executed for $M$ iterations, each time with different assignments. In the initial iteration, the output corresponds to the probability prediction from the vision model. In subsequent iterations, it reflects the probability prediction from the fusion model derived from the previous iteration. This iterative process enables the LM to progressively refine and correct the vision model's predictions.

## 4. Experimental settings
### 4.1. Datasets

In this paper, the experiments are carried out using three publicly available datasets: IR-LPR [51], AOLP [52], UFPR-ALPR [28]. An overview of the datasets is presented in Table 1. Sample images from these datasets can also be seen in Figure 4.

**Table 1.** An overview of the datasets used in our experiments.

| Dataset | Year | Images | Resolution | LP Layout | Evaluation Protocol |
|---|---|---|---|---|---|
| IR-LPR | 2022 | 20967 car images<br>48712 LP images | 1280 × 1280 | Iranian | Yes |
| UFPR-ALPR | 2018 | 4500 car images | 1920 × 1080 | Brazilian | Yes |
| AOLP | 2013 | 2049 car images | Various | Taiwanese | No |

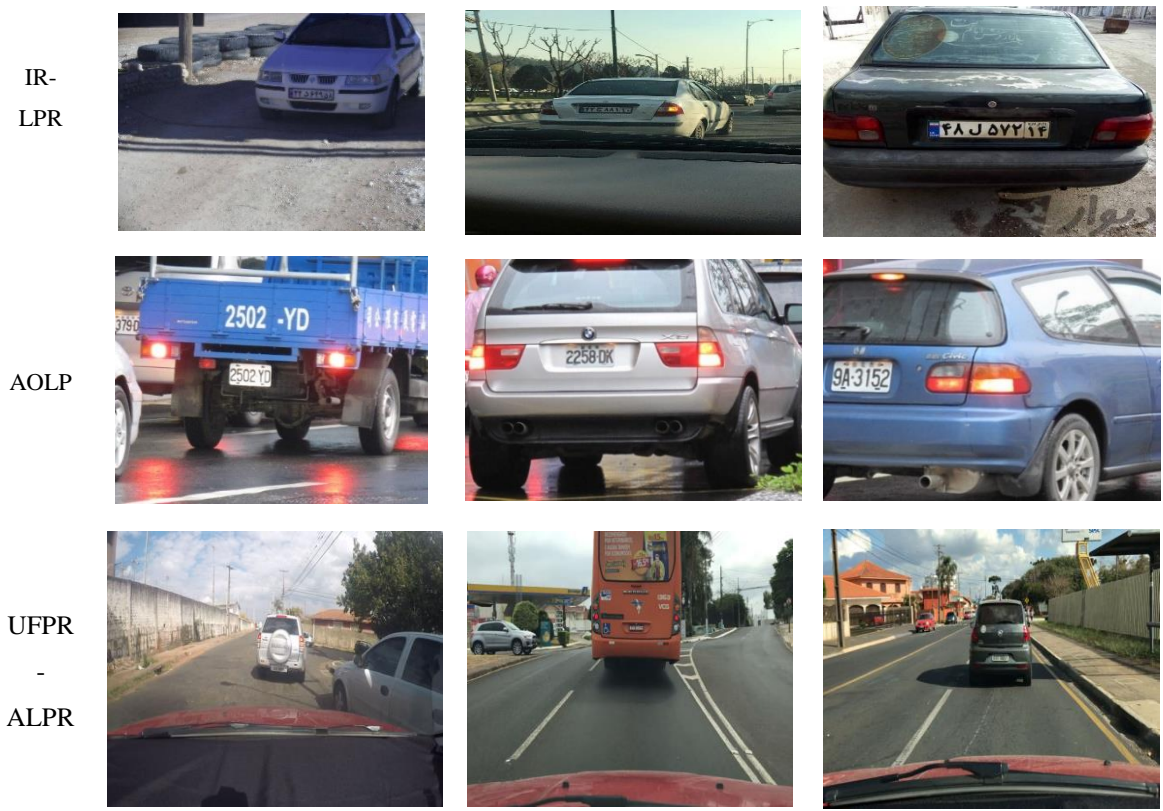

**Fig.4.** Sample images from different datasets.

### 4.1.1. IR-LPR dataset

The IR-LPR database has been created for various purposes, including license plate localization and character recognition (CR) [51].

This database comprises three sets of images:

First Set: Includes 20,967 scene images containing one or more moving vehicles. These images were collected from diverse environments such as multiple car sales websites, parking lots, and at different times of the day under varying lighting conditions. The distances at which these images were captured range from 1 meter to over 10 meters. Most of the images were taken using mobile phone cameras with different resolutions and formats.

Second and Third Sets: Contain only license plate images, with a total of 48,712 images. The primary difference between these two sets is that the third set contains synthetic images, where license plates were artificially generated by placing numbers and letters on license plate backgrounds. These extra images are also used for training and evaluating text recognition step.

This database accounts for various factors such as weather and lighting conditions, camera-to-vehicle distances, license plate viewing angles, and the use of cameras with different quality levels. However, it should be noted that the license plate labels in the database were prone to significant errors. In this study, all license plate annotations were thoroughly reviewed and corrected.

**4.1.2. UFPR-ALPR dataset**

A Brazilian dataset, titled "Federal University of Parana-Automatic License Plate Recognition" (UFPR-ALPR) [50] that includes data from 300 vehicles. The images, each with a resolution of 1920 × 1080 pixels, feature multiple captures of the same vehicle from another moving vehicle. This dataset, comprising approximately 4,500 images, is particularly valuable for developing license plate recognition systems intended for deployment in police vehicles. Notably, the dimensions of the images vary since they focus solely on license plates. The UFPR-ALPR dataset is challenging due to the use of three different non-static cameras, which captured images of various vehicle types (cars, motorcycles, buses, and trucks) within complex backgrounds and under diverse lighting conditions [22].

**4.1.3. AOLP Dataset**

This dataset comprises 2,049 vehicle images designed for evaluating license plate recognition systems. It is divided into three main subsets:

AC Subset: Consists of 681 vehicle images captured under controlled conditions, such as at parking entrances and exits.

LE Subset: Includes 757 images of vehicles moving on roads and highways, intended for surveillance and law enforcement applications.

RP Subset: Contains 611 vehicle images collected to simulate roadside patrol conditions.

Each subset features diverse images in terms of viewing angles, camera-to-vehicle distances, and varying weather and lighting conditions. All images are meticulously annotated, with license plate positions and contents precisely labeled. This makes the dataset highly suitable for the development and evaluation of license plate detection and character recognition algorithms.

## 4.2. Training Settings

The configuration of the model for license plate detection and recognition involves adjusting several critical hyperparameters that influence the model's training process and overall performance. Among these, the image size determines the resolution of the input fed into the model, directly affecting the quality of feature extraction. The learning rate is another essential hyperparameter, controlling how quickly the model updates its weights during training. Its optimal value depends on the complexity of the task and the characteristics of the dataset. Additionally, the batch size, which defines the number of images processed simultaneously in each training iteration, is configured based on the model's complexity to ensure efficient use of computational resources.

The experiments are conducted on a computer equipped with an Intel(R) Core(TM) i9-10900k @ 3.70GHz processor, 32GB of RAM, a 1024GB SSD, and an NVIDIA Geforce RTX 3070 GPU with 8GB of GDDR6 memory, providing a robust computational setup for effective training.

### 4.3. Evaluation Metrics

**4.3.1. License Plate Detection Metrics**

For the validation of object detection models, common metrics such as Precision, Recall, Average Precision (AP), Mean Average Precision (mAP), and F1-score are used [53], which will be briefly explained in the following.

**Precision:** The Precision criterion measures the proportion of positively predicted samples that are accurately identified and is calculated using equation (8).

$$precision = \frac{TP}{TP + FP} \qquad (8)$$

True Positive (TP) means that the model has identified an object in a specific situation and this decision was correct, and False Positive (FP) means that the model has identified an object in a specific situation and this decision was wrong.

**Recall:** The Recall criterion measures the proportion of positive samples that were correctly predicted and is calculated using equation (9).

$$Recall = \frac{TP}{TP + FN} \qquad (9)$$

A false negative (FN) occurs when the model fails to recognize an object in a specific situation, resulting in an incorrect decision.

**Mean Average Precision (mAP):** In multi-class problems, the validity of the model can be measured by the average accuracy of the classes. However, this criterion does not provide information about the details, accuracy, and calling of each class, which is considered a weakness and is not enough to validate the model.

**F1-score:** It measures the balance between accuracy and recall. When the value of F1 is high, it means that the value of accuracy and recall is high at the same time. F1 score is calculated from equation (10).

$$F1\ Score = 2 * \frac{Precision * Recall}{Precision + Recall} \tag{10}$$

### 4.3.2. License Plate Recognition Metric

In recognition systems, evaluation is a critical phase as it determines the system's performance and accuracy. In this study, accuracy is utilized as one of the key metrics for evaluating the proposed methods. This metric represents the percentage of license plates that are correctly recognized [54]. Essentially, it reflects the system's ability to accurately identify the characters present on the plates.

Accuracy is influenced by factors such as the font type used, the license plate design, and the image quality. Accurate character recognition is crucial for the optimal performance of an LPR system, as a single misrecognized character can lead to incorrect plate identification. The equation used to calculate this metric is as follows:

$$Acc = \frac{The\ number\ of\ recognized\ license\ plates.}{The\ total\ number\ of\ license\ plates} \tag{11}$$

## 5. Results

In this section, we present the experiments conducted to evaluate the effectiveness of the proposed ALPR system. We begin by individually assessing the detection stages, as the regions used in the license plate (LP) recognition stage are derived from detection results rather than being cropped directly from the ground truth.

To train a model, the images in the dataset must be divided into training, validation, and test sets. Table 4 outlines the structure of these sets for the three datasets used in this study. As noted in Table 1 the IR-LPR and UFPR-ALPR datasets include an evaluation protocol. This means that in the original references for these datasets, the images are pre-divided into training, validation, and test sets, and the experiments are conducted using the configurations specified in those references.

In contrast, the AOLP dataset does not provide such a predefined division. However, consistent with other studies, the images in this research are randomly split into training, validation, and test sets, as described in Table 3.

**Table 3.** The number of images for training, validation, and test sets across the three datasets.

| Dataset | IR-LPR | | UFPR- ALPR | | AOLP | |
|---|---|---|---|---|---|---|
| | detection | Recognition | detection | recognition | Detection | Recognition |
| Training Samples | 14670 | 34098 | 1800 | 1800 | 1126 | 1126 |
| Validation Samples | 4177 | 7308 | 900 | 900 | 225 | 225 |
| Test Samples | 2120 | 7306 | 1800 | 1800 | 698 | 698 |
| Total | 20967 | 48712 | 4500 | 4500 | 2049 | 2049 |

## 5.1. License Plate Detection Results

The experimental results of comparing YOLO, which was used as the license plate detection algorithm in this paper with the state-of-the-art approaches on the three datasets, are shown in Table 4. Figure 5 presents some qualitative results using the license plate detection system. As this figure shows, bounding boxes related to license plates are accurately detected in many images. While there are other boxes very close to license plates in the scene.

**Table 4.** Detection outcomes obtained across all datasets. Tr: train, Val: validation and Te:test.

| Dataset | Precision (%) | | | Recall (%) | | | F-Score | | | mAP@0.5 | | |
|---|---|---|---|---|---|---|---|---|---|---|---|---|
| | Tr | Val | Te | Tr | Val | Te | Tr | Val | Te | Tr | Val | Te |
| IR-LPR | 100 | 100 | 100 | 97 | 97 | 97 | 98.48 | 98.48 | 98.48 | 97.2 | 96.7 | 97.4 |
| UFPR-ALPR | 100 | 100 | 100 | 100 | 99 | 100 | 100 | 99.5 | 100 | 99.4 | 98 | 98.5 |
| AOLP | 100 | 100 | 100 | 100 | 100 | 100 | 100 | 100 | 100 | 99.1 | 99.4 | 99.1 |

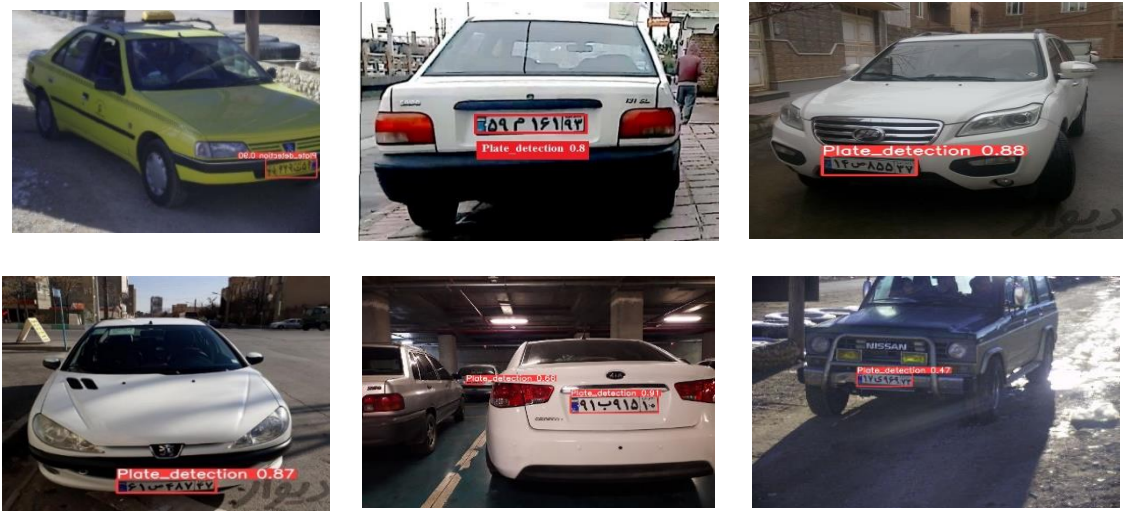

(a) IR-LPR dataset

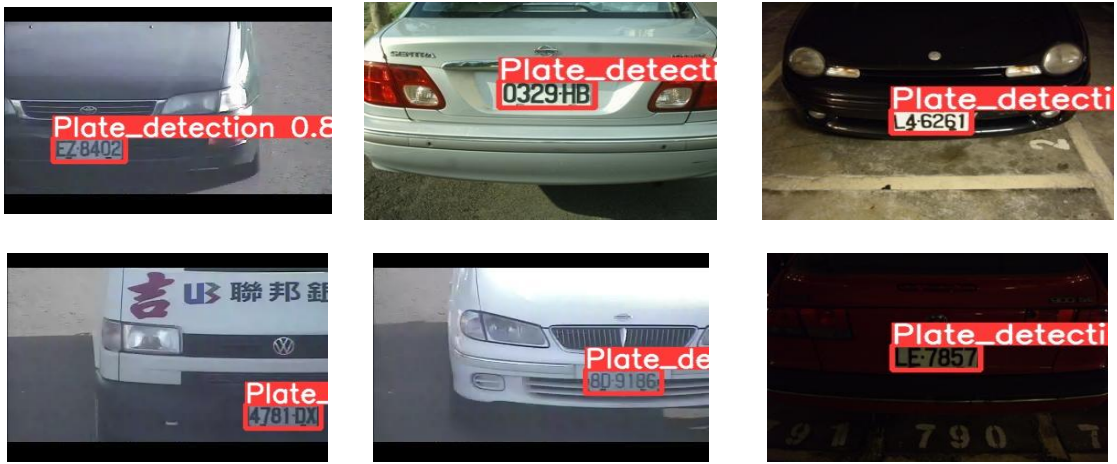

(b) AOLP dataset

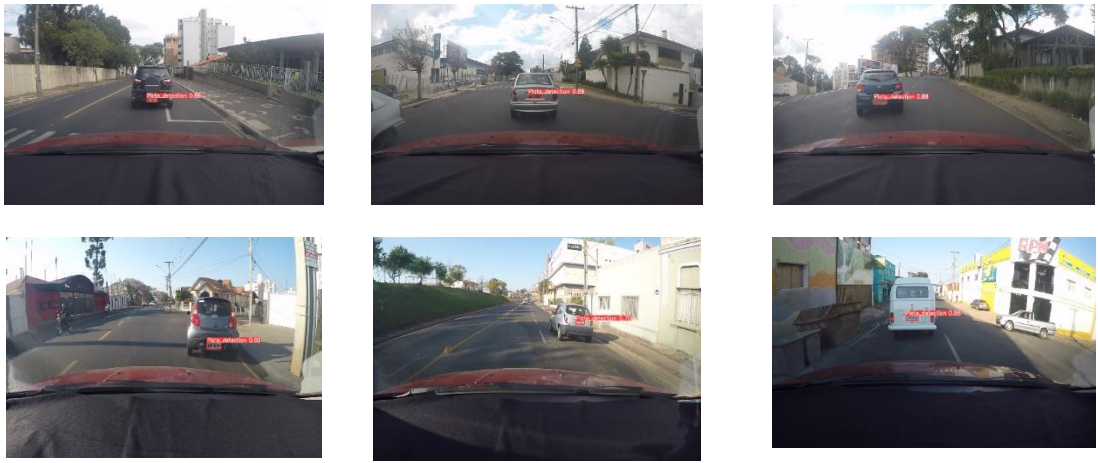

(c) UFPR-ALPR dataset

**Fig. 5.** Results of LP detection.

## 5.2. License Plate Recognition Results

Table 5 presents the results of the proposed model during the training, evaluation, and testing phases for each of the three datasets: IR-LPR, AOLP, and UFPR-ALPR. As shown in the table, training, evaluation, and testing were performed with good accuracy for all three datasets.

**Table 5.** Recognition accuracy of the recognition model for different datasets.

| Dataset | Train | Validation | Test |
|---|---|---|---|
| IR-LPR | 99.97 | 97.03 | 97.12 |
| UFPR-ALPR | 99.99 | 99.9 | 99.93 |
| AOLP | 100 | 99.99 | 99.4 |

Figure 6 presents qualitative results for several images from the three datasets along with the corresponding recognition outputs produced by the proposed model. These results demonstrate that the proposed model is capable of recognizing many license plates with very high accuracy. However, for certain plates, even incorporating the language model could not improve the results. For instance, in the last row of the figure, a license plate from the AOLP dataset is shown where the number "8" has been misrecognized as the letter "B". Unfortunately, since in this dataset both letters and numbers may appear without a clear rule governing their usage, the model is unable to perform accurate recognition in this case.

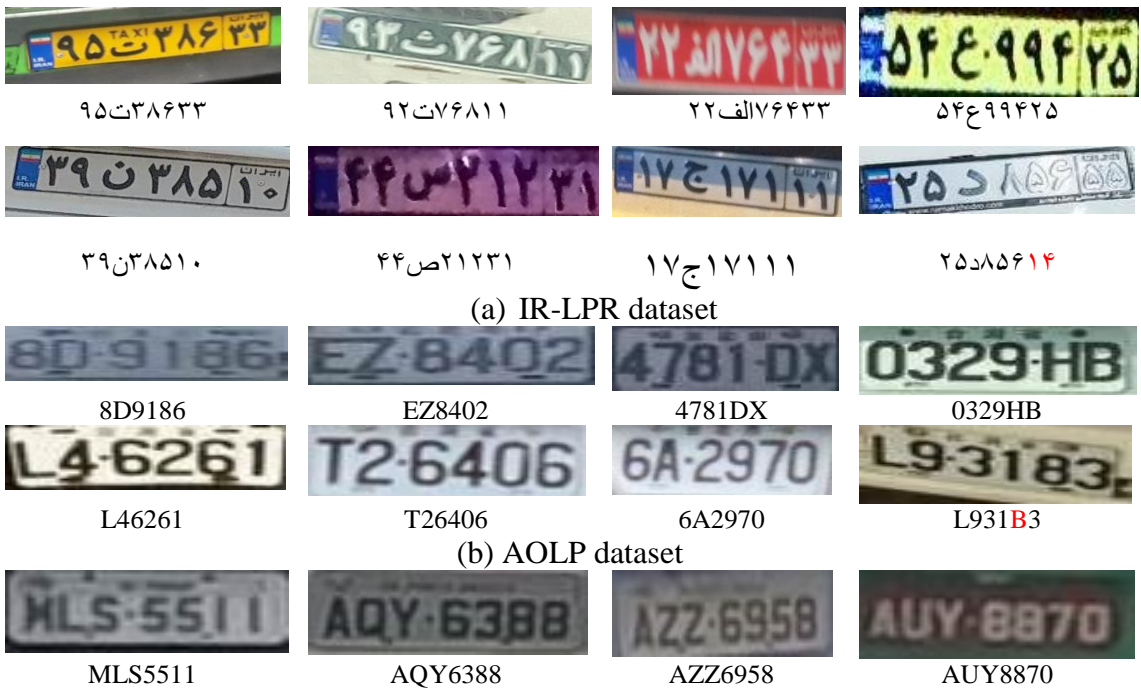

(a) IR-LPR dataset

(b) AOLP dataset

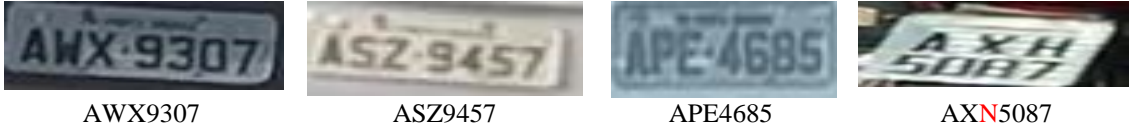
AWX9307  ASZ9457  APE4685  AXN5087

(c) UFPR-ALPR dataset

**Fig. 6.** The results of license plate recognition using for some sample images.

### 5.3. End-to-End Recognition Accuracy

The end-to-end performance of the proposed system is used for assessing its practical usability in real-world scenarios. This performance hinges on two key stages: the License Plate Detection (LPD) stage and the Character Recognition (CR) stage. Failures in either stage directly impact the overall system accuracy. Specifically, if the LPD stage does not successfully detect all license plates in an image or if the CR stage misclassifies characters compared to ground-truth labels, the end-to-end accuracy will degrade.

Results presented in Table 6 show the end-to-end accuracy of proposed method for different datasets. Even when tested with skewed plates or non-uniform illumination, the system effectively detected and cropped license plates and accurately recognized their characters.

**Table 6 .** End to End accuracy of proposed method for different datasets.

| Dataset | Accuracy |
|---|---|
| IR-LPR | 94.77 |
| AOLP | 97.56 |
| UFPR-ALPR | 99.99 |

### 5.4. Computational Efficiency

The proposed method integrates detection and recognition models to deliver a comprehensive solution. Together, these models balance speed and precision, making them suitable for real-time applications.

The combined model achieves an average total processing time of 55.565 milliseconds per image (detection and recognition). Table 7 provides a consolidated overview of the parameters, computational requirements, and inference speed for the integrated system. Despite the combined computational demands of 198.0 GFLOPs and $95 \times 10^4$ parameters, the system remains efficient and practical for real-time deployment scenarios.

**Table 7 .** Parameters and Computational Requirements of the Integrated Detection and Recognition System.

| Component | Inference | FLOPs | Parameters |
|---|---|---|---|

|  | Speed (ms/image) | (*10$^9$) | (*10$^6$) |
| --- | --- | --- | --- |
| Detection | 23.4 | 190.8 | 58 |
| Recognition | 32.165 | 7.2 | 37 |
| Total | 55.565 | 198.0 | 95 |

## 5.5. Comparison with other state-of-the-art methods

Table 8 shows the results of the proposed model in comparison with other methods. According to the results presented in Table 9, the proposed model demonstrates higher recognition accuracy on all three datasets: AOLP, UFPR-ALPR, and IR-LPR, compared to other methods.

- On the AOLP dataset, the proposed model achieved a recognition accuracy of 99.99%, which is 0.2% higher than the best result achieved by previous methods (99.7%).

- On the UFPR-ALPR dataset, the proposed model also outperformed other methods with an accuracy of 99.93%, showing a significant difference in performance.

- On the IR-LPR dataset, the proposed model achieved a recognition accuracy of 97.12%, significantly outperforming other models.

It should be noted that the first two datasets have a limited number of samples, which may not fully demonstrate the power of the proposed model in comparison to others. Despite the superior performance of the proposed model, other methods also perform acceptably well on these limited samples. However, for the IR-LPR dataset, which includes a much larger and more diverse set of samples, the difference in performance is more pronounced.

The comparison with two vision-based methods [55] and [39] shows that the use of a language model alongside a vision model has contributed to improving the recognition accuracy.

This performance advantage can be attributed to the combined use of vision and language models in the data processing and feature extraction stages, especially with the use of advanced language models. This approach allows the system to better simulate the complex information structures in license plates, leading to not only improved recognition but also higher accuracy in simulating words and numbers in challenging environments.

Overall, the proposed model has managed to improve recognition accuracy across all the datasets used. These results indicate the stronger performance of the proposed model compared to other similar methods and competitors, particularly in non-ideal conditions and low-quality images, which typically lead to a decline in accuracy in other models.

**Table 8.** Comparison of the results of different license plate detection methods.

| Dataset | IR-LPR | | | AOLP | | | UFPR-ALPR | | |
|---|---|---|---|---|---|---|---|---|---|
| | P | R | F1 | P | R | F1 | P | R | F1 |
| Hao et al.2024 [55] | 94.9 | 60.1 | 73.6 | | | | | | |
| Rahmati et al.2023 [39] | 96.2 | 95.7 | - | | | | | | |
| Sarfraz et al.2003 [56] | | | | 91 | 95 | - | | | |
| Selmi et al al.2017 [54] | | | | 93.5 | 93 | - | | | |
| Li et al.2016 [57] | | | | 97.7 | 97.6 | - | | | |
| Li et al.2018 [58] | | | | - | 99.3 | - | | | |
| Selmi et al.2020 [59] | | | | 99.2 | 99.2 | - | | | |
| Pham et al.2023 [60] | | | | 98.3 | 99 | 98.65 | | | |
| Laroca et al.2021 [17] | | | | - | 99.85 | - | 99.57 | 100 | - |
| Kesentini et al.2019 [61] | | | | | | | 75 | 98 | 85 |
| Wang et al.2019 [62] | | | | | | | 98 | 92 | 95 |
| Chowdhury et al. 2020 [63] | | | | | | | 99 | 95 | 95 |
| Redmon et al.2017 [25] | | | | | | | 92.7 | 94.7 | 93.7 |
| Redmon et al.2018 [64] | | | | | | | 94.9 | 97.4 | 96.1 |
| Bochkovskiy et al.2020 [65] | | | | | | | 93.1 | 92.6 | 92.8 |
| Ding et al.2024 [66] | | | | | | | 98.3 | 98.7 | **98.5** |
| Proposed Method | **100** | **97** | **98.48** | **100** | **100** | **100** | **100** | **100** | **100** |

**Table 9.** Performance comparison of accuracy of different license plate recognition methods.

| Dataset | IR-LPR | AOLP | UFPR-ALPR |
|---|---|---|---|
| Hao et al.2024 [55] | 94.9 | | |
| Laroca et al.2021 [17] | | 99.2 | |
| Silva et al.2018 [39] | | 98.36 | |
| Pham et al.2023 [60] | | 98.1 | |
| Laroca et al.2019 [17] | | | 97.57 |
| Fernandes et al.2020 [67] | | | 96.87 |
| Masood et al.2017 [68] | | | 76.7 |
| Ding et al.2024 [66] | | | 90 |
| Proposed Method | **97.12** | **99.4** | **99.93** |

Table 10 presents the end-to-end accuracy of proposed method compared with other methods. As this table shows, the results are comparable with state-of-the-art methods.

**Table 10.** Comparison of various methods for end-to-end recognition accuracy.

| Dataset | AOLP | UFPR-ALPR | IR-LPR |
|---|---|---|---|
| AI-Batat et al. 2022 [69] | **98** | 73.3 | |
| Laroca et al. 2021 [17] | | 90.0 | |
| Proposed Method | 97.56 | **99.99** | **94.77** |

## 5.6. Nighttime license plate recognition

One of the challenges faced by license plate detection and recognition systems is handling nighttime images, which typically involve greater difficulty due to low lighting conditions. In this section, the IR-LPR license plate test set was divided into two subsets: daytime and nighttime images. The evaluation performed on 889 nighttime license plate images achieved an end-to-end accuracy of 94.60%. These results indicate that the recognition accuracy is not only satisfactory for the entire dataset but also reasonably high for nighttime images. Figure 7 illustrates some qualitative results for nighttime images. As shown, many of these images suffer from issues such as improper angles, poor lighting conditions, shadows, or long distances. Despite these challenges, the recognition accuracy remains acceptable.

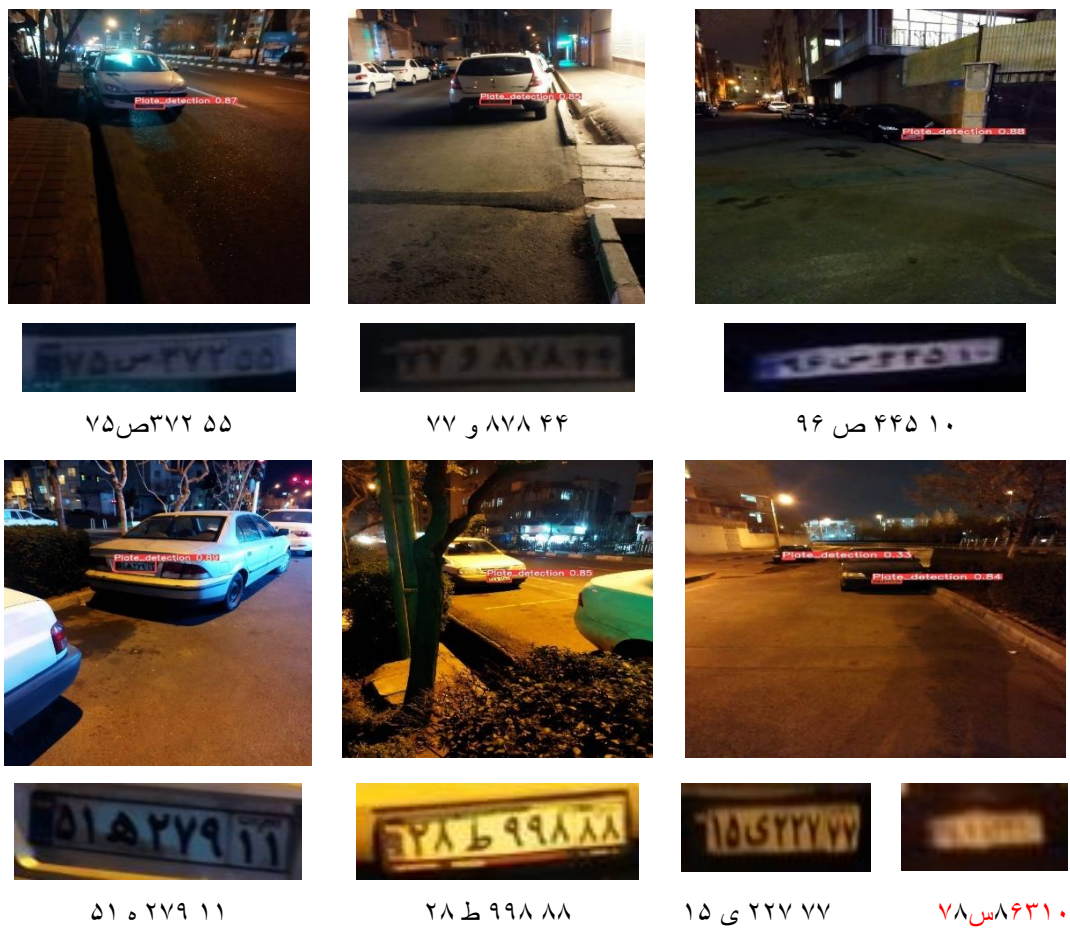

**Fig.7.** Some qualitative results of night images.

## 5.7. Analysis and Discussion

The experimental results comprehensively validate the efficacy of our proposed pattern-aware vision–language framework for Automatic License Plate Recognition. The key finding is the system's ability to maintain high accuracy and robustness across diverse plate layouts and challenging real-world conditions, without the need for manual feature engineering or layout-specific heuristic rules.

The superior detection rates presented in Table 4, achieved by employing a modern, high-precision object detector, confirm the system's robust ability to localize the license plate region under varying lighting conditions and visual noise. Crucially, by feeding this detected region directly into the recognition stage, our system successfully eliminates the character segmentation bottleneck, which is a common source of accumulated error in conventional multi-stage ALPR pipelines.

Table 5, which details the recognition performance, is central to our claim of layout-independence. The integrated vision and language models are shown to effectively learn the intricate structural patterns and formatting constraints specific to license plates. Unlike previous methods relying on explicit post-OCR correction rules for specific layouts (e.g., US, UK), our system autonomously refines its output by embedding pattern analysis within the recognition loop itself. This is demonstrated by the consistently high recognition accuracy achieved across the three diverse international datasets: IR-LPR, UFPR-ALPR, and AOLP, each featuring unique character sets and structural arrangements.

Tables 8, 9, and 10 highlight the system's overall superior performance in comparison to leading segmentation-free approaches. This edge in both recognition and end-to-end accuracy confirms the effectiveness of the vision–language integration. Specifically, the system's iterative refinement mechanism, driven by the language model, leverages contextual cues to correct visual errors (e.g., misclassifying 'G' and '6' or 'I' and '1'), simulating a human-like understanding of the expected plate pattern. The significant gains observed, particularly on the IR-LPR dataset which provides a broad and challenging test of generalizability, underscore the success of the pattern-aware approach.

Furthermore, the overall end-to-end accuracy (Table 6) shows only a minor degradation compared to the standalone character recognition accuracy, indicating the stability of the unified pipeline and the limited propagation of errors from the detection stage.

Finally, the system's robustness is further emphasized by its performance in challenging low-light conditions, achieving an impressive end-to-end accuracy of 94.60% on the IR-LPR nighttime subset. Coupled with the efficient average processing time of 55.565 milliseconds per image (Table 7), the proposed framework is proven to be a robust, efficient, and highly adaptable solution well-suited for high-demand intelligent transportation applications.

# 6. Conclusion

In this paper, we presented a novel pattern-aware vision–language framework for Automatic License Plate Recognition that fundamentally addresses the limitations of multi-stage, layout-dependent systems. We successfully unified the ALPR pipeline by integrating a high-precision object detector with a singular recognition stage based on a convolutional transformer and an iterative language model. This design choice effectively eliminates the traditional character segmentation module, a critical bottleneck prone to error propagation.

The core contribution of this work lies in its layout-independent mechanism. By embedding pattern analysis directly into the recognition loop, the framework autonomously learns the structural and formatting constraints of license plates from the data itself. This eliminates the reliance on rigid, rule-based post-OCR corrections and enables inherent adaptability to new plate designs and languages. Experimental validation across the IR-LPR, UFPR-ALPR, and AOLP datasets confirmed the superior performance and robustness of our approach, showcasing its stability under adverse conditions and its scalability across diverse international layouts.

For future research, we aim to extend the framework's application to full video-based ALPR systems, focusing on maintaining its real-time efficiency on constrained edge devices. Furthermore, optimizing the language model component to simultaneously process and validate highly complex, multi-script license plates (such as those containing both Latin and Farsi characters) remains a key direction for enhancing global applicability.

## Data Availability

This study utilized publicly available datasets, which have been appropriately cited in the manuscript. No new data were generated or analyzed during this research.

## Refrences